\documentclass[10pt, conference, compsocconf]{IEEEtran}
\ifCLASSINFOpdf
\else
\fi

\usepackage{amsmath}
\usepackage{amsfonts}
\usepackage{graphicx}
\makeatletter
\newcommand{\linebreakand}{%
  \end{@IEEEauthorhalign}
  \hfill\mbox{}\par
  \mbox{}\hfill\begin{@IEEEauthorhalign}
}
\makeatother

\begin{document}
\bibliographystyle{unsrt}
%
\title{Anomaly Detection with Adversarially Learned Perturbations of Latent Space}


\author{\IEEEauthorblockN{Vahid Reza Khazaie}
\IEEEauthorblockA{Department of Computer Science\\
University of Western Ontario\\
London, Canada\\
vkhazaie@uwo.ca}
\and

\IEEEauthorblockN{Anthony Wong}
\IEEEauthorblockA{Department of Computer Science\\
University of Western Ontario\\
London, Canada\\
awong655@uwo.ca}
\and

\IEEEauthorblockN{John Taylor Jewell}
\IEEEauthorblockA{Department of Computer Science\\
University of Western Ontario\\
London, Canada\\
jjewell6@uwo.ca}
\and 

\linebreakand 
\IEEEauthorblockN{Yalda Mohsenzadeh}
\IEEEauthorblockA{Department of Computer Science\\
University of Western Ontario\\
London, Canada\\
ymohsenz@uwo.ca}
}


%


\maketitle

\begin{abstract}
Anomaly detection is to identify samples that do not conform to the distribution of the normal data. Due to the unavailability of anomalous data, training a supervised deep neural network is a cumbersome task. As such, unsupervised methods are preferred as a common approach to solve this task. Deep autoencoders have been broadly adopted as a base of many unsupervised anomaly detection methods. However, a notable shortcoming of deep autoencoders is that they provide insufficient representations for anomaly detection by generalizing to reconstruct outliers. In this work, we have designed an adversarial framework consisting of two competing components, an Adversarial Distorter, and an Autoencoder. The Adversarial Distorter is a convolutional encoder that learns to produce effective perturbations and the autoencoder is a deep convolutional neural network that aims to reconstruct the images from the perturbed latent feature space. The networks are trained with opposing goals in which the Adversarial Distorter produces perturbations that are applied to the encoder's latent feature space to maximize the reconstruction error and the autoencoder tries to neutralize the effect of these perturbations to minimize it. When applied to anomaly detection, the proposed method learns semantically richer representations due to applying perturbations to the feature space. The proposed method outperforms the existing state-of-the-art methods in anomaly detection on image and video datasets.

\end{abstract}


%
\IEEEpeerreviewmaketitle

\section{Introduction}
Detecting abnormal samples from a group of normal data is the goal of anomaly detection. Anomaly detection differs from common supervised classification problems due to either poor sampling or inaccessibility of abnormal data during training. Therefore, one-class classification is an efficient approach for solving this task.

In computer vision, anomaly detection has widespread applications such as finding biomedical markers \cite{schlegl2017unsupervised} and monitoring video surveillance footage \cite{luo2017revisit}. The task of detecting anomalies in images involves identifying whether an image is an inlier or an outlier based on training data that primarily consists of inlier images. As a solution to the lack of outlier samples, one-class classification methods attempt to model the distribution of only the inlier data \cite{zimek2012survey}. A sample that does not match the inlier distribution is considered an outlier. The high dimensionality in which the data points exist makes it difficult to model the distribution of image data with conventional methods \cite{zimek2012survey}.

Deep learning has contributed to developing methods that effectively produce representations of high-dimensional data \cite{bengio2013representation}. Of these methods, Autoencoders (AEs) are an unsupervised class of algorithms that are suitable for modeling image data \cite{bengio2007greedy}. A standard AE consists of two components: an encoder and a decoder. Encoders learn to map images into a latent space, while decoders learn mappings from the latent space to original images. The model weights are optimised by minimizing the error between the original image (input to the encoder) and its reconstruction (output of the decoder).

In many approaches to one-class classification, AEs serve as a powerful unsupervised way to learn representations for anomaly detection \cite{chalapathy2019deep}. Prior to detecting abnormal images, the AE is trained on sets of mostly normal images. An anomaly score is calculated using the reconstruction error of a sample. As a result, the reconstruction error is expected to be lower for inliers compared to outliers \cite{xia2015learning}. However, this assumption is not always true, and the AE can reconstruct images outside the distribution of the training data as well \cite{zong2018deep, gong2019memorizing}. This is particularly evident when abnormal images share patterns with inliers.

In more recent methods, the various modifications to the autoencoder architecture has resulted in higher anomaly detection performance \cite{abati2019latent, perera2019ocgan, zimmerer2018context}. In the Denoising Autoencoders (DAE) \cite{vincent2010stacked}, the model learns to remove added noise from an input. Denoising has also been demonstrated to provide more robust representations. A specific type of DAE, context autoencoders (CAE) \cite{pathak2016context}, has demonstrated excellent performance in anomaly detection. In contrast to adding noise, random masking is applied to input images, and the reconstruction task involves inpainting the randomly masked region. In this way, random masking implicitly forces CAEs to learn semantic information about the distribution of training data \cite{pathak2016context}. In some cases, CAEs have difficulties detecting anomalies because of suboptimal representations.

In relation to anomaly detection, some deep autoencoders generalize well to outliers causing them to fail at detecting anomalies. To alleviate this issue, we designed an adversarial framework consisting of two competing components, an Adversarial Distorter, and an Autoencoder. The adversarial distortor aims to maximize the encoder's reconstruction error by applying perturbations to the latent space, while the autoencoder attempts to minimize it by neutralizing the effects of these perturbations. Contrary to previous AE-based approaches, our framework enables autoencoders to learn semantically richer representations by overcoming the added perturbation. In this paper, our contributions are the following:

\begin{itemize}
    \item We propose a novel method of representation learning to improve the performance of autoencoders for anomaly detection in images and videos.
    \item Our framework is optimized in an adversarial setting which yields more efficient representations.
    \item With the learned perturbations added to the latent space of the autoencoder, the reconstruction loss for anomalies will be higher which improves the anomaly detection performance.
\end{itemize}




\section{Related Works}
Novelty, outliers, and anomalies are commonly detected using one-class classification \cite{sakurada2014anomaly, zhai2016deep, zhou2017anomaly, zong2018deep, chong2017abnormal, jewell2021oled, scholkopf2002learning, perera2019ocgan, sabokrou2018adversarially, zaheer2020old}. This problem is solved by trying to identify outliers or novel concepts by capturing the distribution of the inlier class. Traditional tools for identifying such distributions include support vector machines (SVMs) \cite{scholkopf2002learning, hayton2000support} and principal component analysis (PCA) \cite{bishop2006pattern, hoffmann2007kernel}. There are also unsupervised clustering methods like k-means \cite{zimek2012survey} and Gaussian Mixture Models (GMM) \cite{xiong2011group} that have been used for detecting anomalies among normal data distributions, but these methods fail to handle high-dimensional data. 
  
Many proposed methods utilize reconstruction error to determine whether a particular sample conforms to the distribution of training data \cite{xu2015learning, sabokrou2016video}. Moreover, many deep learning-based models with an autoencoder architecture \cite{sakurada2014anomaly, zhai2016deep, zhou2017anomaly, zong2018deep, chong2017abnormal} also detect anomalies based on reconstruction error. But in some cases, these methods have been shown to generalize well and can reconstruct out-of-distribution samples, therefore they fail to detect anomalies at the test stage.

\begin{figure*}[ht]
\begin{center}
   \includegraphics[width=7in]{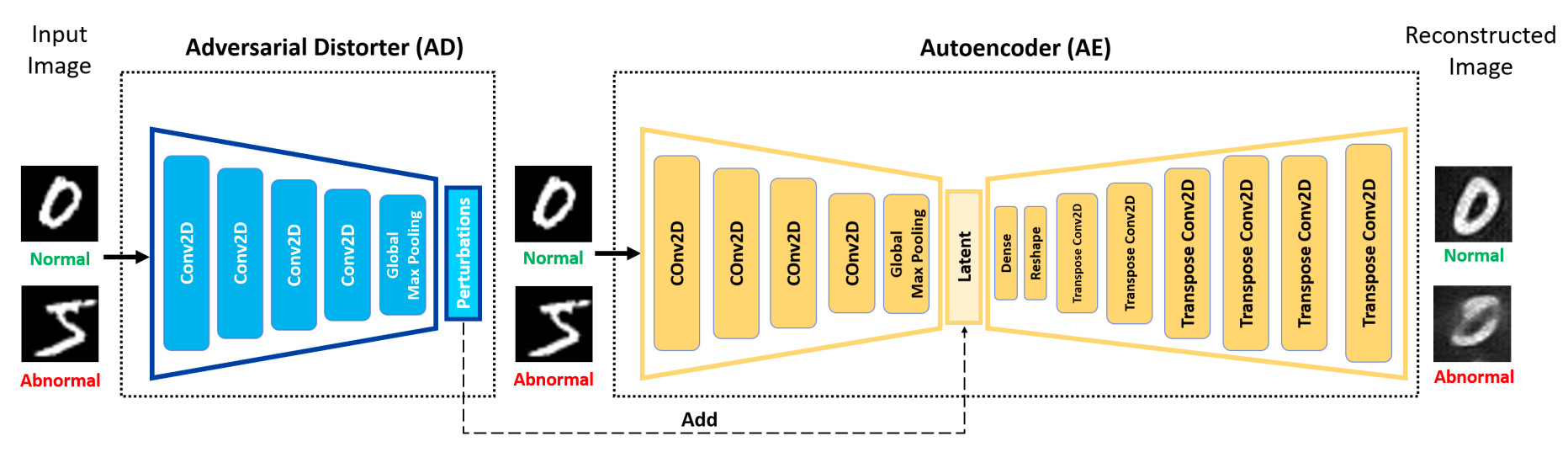}
\end{center}
   \caption{Overview of ALPS architecture. From the input, the Adversarial Distorter learns how to produce perturbations. The Autoencoder minimizes reconstruction error, while the Adversarial Distorter maximizes it.}
\label{modelarch}
\end{figure*}

GANs \cite{goodfellow2014generative} are generative models that are composed of two neural networks - a Generator (G) and a Discriminator (D). From a random noise vector $z$, G is supposed to create realistic-looking images. In D, we are trying to distinguish real images from images that are created by G, which we call fake images. Specifically, D outputs a high score for real images and a low score for fake images to accomplish this objective. Objective functions of GANs are as follows:

\begin{equation}
\begin{aligned}
\min_G \max_D
\mathbb{E}_{x \sim p_{\text{data}}(x)}[\log D(x)] + \\
\mathbb{E}_{z \sim p_{z}(z)}[\log (1 - D(G(z))]
\end{aligned}
\end{equation}

Several works mapped images to a random distribution from an image space using deep convolutional generative adversarial networks (DCGANs) \cite{radford2015unsupervised} to learn a manifold of normal images for anomaly detection \cite{schlegl2017unsupervised, schlegl2019f} (AnoGAN, f-AnoGAN). In \cite{sabokrou2018adversarially}, the authors presented a one-class classification framework consisting of a Reconstructor (R) as a denoising autoencoder and Discriminator (D) as the detector. The two networks are trained end-to-end in an adversarial manner. The same model was extended by Zaheer et al. \cite{zaheer2020old}, who changed the discriminator to classify good and bad reconstructions, thus improving the performance of adversarial classifiers. The autoencoder architecture was used in \cite{perera2019ocgan} to enforce a uniform distribution of the normal instances across the latent space (OCGAN). By masking the input intelligently, \cite{jewell2021oled} learned more robust representations through an adversarial setup.

Using a small student model learning from a large teacher model, \cite{salehi2021multiresolution} and \cite{georgescu2021anomaly} have attempted to exploit deep pretrained networks. The authors of \cite{salehi2021multiresolution} used a VGG-16 \cite{simonyan2014very} to determine the anomaly score by calculating a multilevel loss from the different activations. Interpretability algorithms are also included in their framework to perform anomaly segmentation and locate anomalous regions. Knowledge distillation can produce high performance anomaly detection, but pretraining on millions of labelled images is required, which may not be applicable to other modalities of data. Knowledge distillation may also be unsuitable in practice because the inference is computationally expensive.

In some cases, anomaly detection methods based on learning the distribution of inliers cannot be applied in practice. Anomalies can be generated alongside normal data during training to turn the problem into a supervised classification task. A GAN can be used to create anomalous data, which turns anomaly detection into a binary classification problem. According to \cite{pourreza2021g2d}, they trained a Wasserstein GAN on normal data and used the generator before convergence had been reached. In this way, the generated data deviates from inliers in a controlled manner. Even though they offer new avenues for detecting anomalies, training a neural network to generate outliers is computationally expensive.

To map the input to a latent space more effectively, Gong et al. proposed a deep autoencoder with a memory module \cite{gong2019memorizing}. Using the latent vector, the decoder will retrieve the memory item that is most relevant for reconstruction. In \cite{park2020learning}, they added a new update system to the memory module for modelling the inliers.

\section{Method}
\subsection{Motivation}
 In previous studies, the reconstruction error of AEs have been shown to be a good metric of whether or not a sample follows the distribution specified in the training examples \cite{xia2015learning}. The main drawback of AE is that they generalize well to outliers and therefore learn insufficient representations for anomaly detection. To overcome this issue, we introduce \textbf{Adversarially Learned Perturbations (ALPS)}, a framework for training AEs which learns more effective representations for anomaly detection. These representations are learned by adding adversarially generated perturbations to the latent space of the autoencoder. An overview of the method is shown in Figure \ref{modelarch}.

\subsection{Components of the Framework}

\subsubsection{Autoencoder}
The Autoencoder is a convolutional neural network that consists of an encoder and a decoder. The encoder maps the input to a latent space through four convolutional layers followed by a global average pooling. The decoder maps from the latent space back to the image, beginning with a dense and a reshaping layer followed by six transpose convolutional layers for upsampling. An autoencoder is trained to remove the perturbations produced by the adversarial distorter. 

\begin{figure*}[ht]
\begin{center}
   \includegraphics[width=7in]{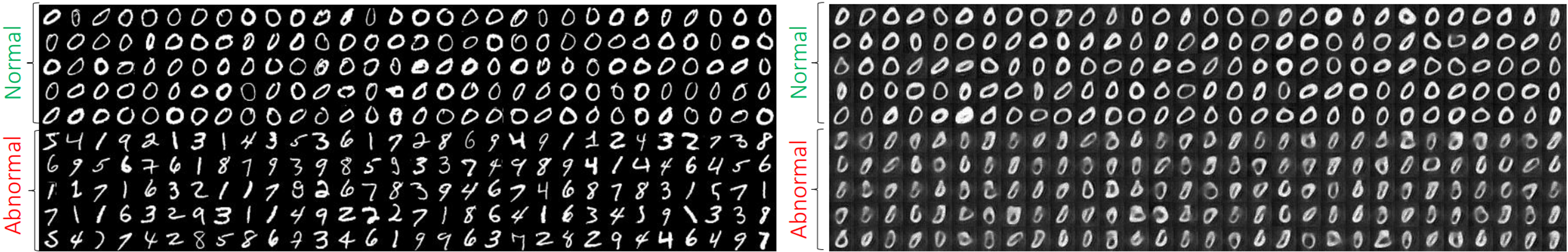}
\end{center}
   \caption{A visual demonstration of the performance of our method on MNIST dataset. The left image is the input to the model and the right image is the reconstruction of the model from the perturbed latent representation.}
\label{mnist_vis}
\end{figure*}

\begin{figure*}[ht]
\begin{center}
   \includegraphics[width=7in]{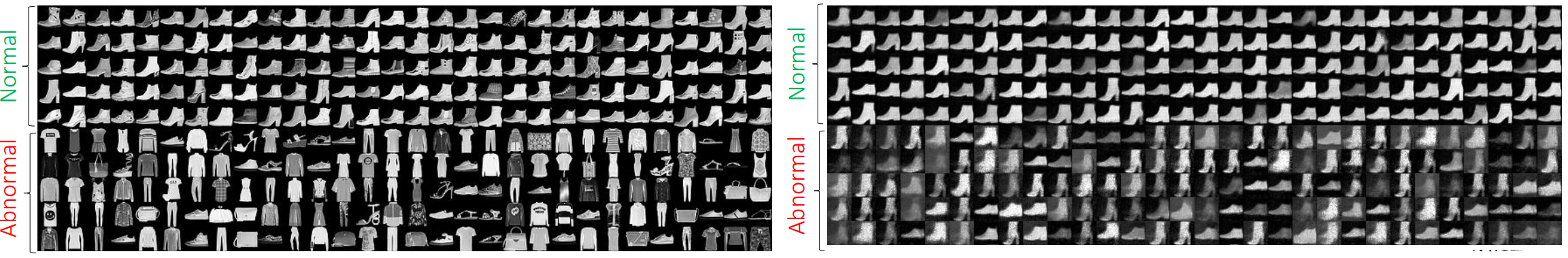}
\end{center}
   \caption{A visual demonstration of the performance of our method on FMNIST dataset. The left image is the input to the model and the right image is the reconstruction of the model from the perturbed latent representation.}
\label{fmnist_vis}
\end{figure*}

\begin{figure*}[ht]
\begin{center}
   \includegraphics[width=7in]{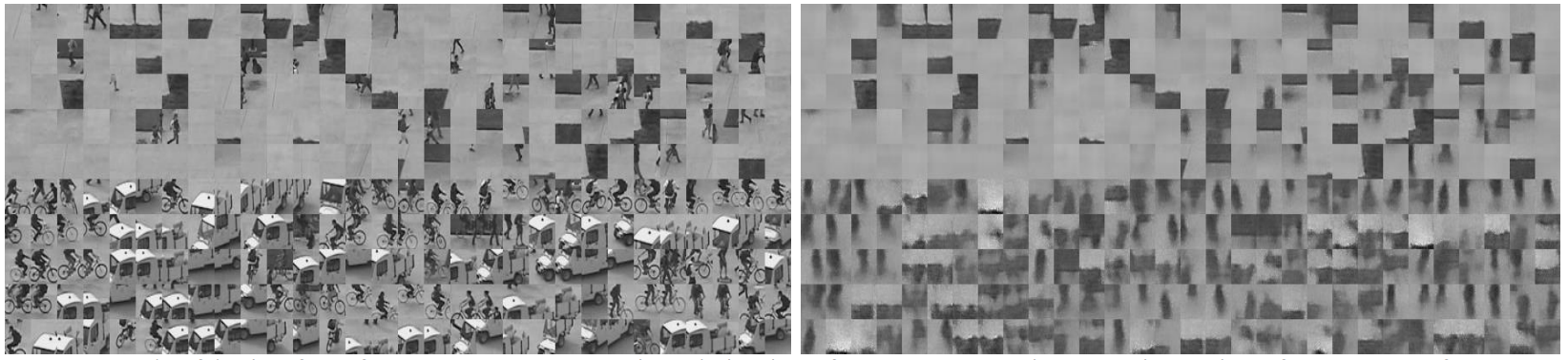}
\end{center}
   \caption{(Left) The first five rows are normal and the last five rows are abnormal patches from UCSD frames. These patches are used as the input to the model. (Right) The first five rows are reconstruction of normal and the last five rows are reconstruction of abnormal patches from UCSD frames.}
\label{ucsd_vis}
\end{figure*}

\subsubsection{Adversarial Distorter}
The Adversarial Distorter is a convolutional encoder that generates perturbations from a given input. The perturbations are used to make reconstruction more difficult for the autoencoder. 

\begin{table*}[htbp]
\centering
\caption{AUROC in \% for anomaly detection on MNIST \cite{lecun1998mnist} and  FMNIST \cite{xiao2017fashion} datasets.}
\label{table:mnist,fnmnist}
\resizebox{\textwidth}{!}{\begin{tabular}{|c|c|c|c|c|c|c|c|c|c|c|c|c|}
\hline
Dataset & Method & 0 & 1 & 2 & 3 & 4 & 5 & 6 & 7 & 8 & 9 & Mean\\

\hline

MNIST
& AnoGAN\cite{schlegl2017unsupervised} & 96.6 & 99.2 & 85.0 & 88.7 & 89.4 & 88.3 & 94.7 & 93.5 & 84.9 & 92.4 & 91.3\\
& DSVDD\cite{ruff2018deep} & 98.0 & 99.7 & 91.7 & 91.9 & 94.9 & 88.5 & 98.3 & 94.6 & 93.9 & 96.5 & 94.8\\
& OCSVM\cite{scholkopf2002learning} & 99.5 & 99.9 & 92.6 & 93.6 & 96.7 & 95.5 & 98.7 & 96.6 & 90.3 & 96.2 & 96.0\\
& CapsNet\textsubscript{PP} \cite{li2020exploring} & 99.8 & 99.0 & 98.4 & 97.6 & 93.5 & 97.0 & 94.2 & 98.7 & 99.3 & 99.0 & 97.7\\
& OCGAN\cite{perera2019ocgan} & 99.8 & 99.9 & 94.2 & 96.3 & 97.5 & 98.0 & 99.1 & 98.1 & 93.9 & 98.1 & 97.5\\
& LSA\cite{abati2019latent} & 99.3 & 99.9 & 95.9 & 96.6 & 95.6 & 96.4 & 99.4 & 98.0 & 95.3 & 98.1 & 97.5\\

& \textbf{Ours (ALPS)} & ${99.68}$ & ${99.92}$ & ${94.09}$ & ${96.19}$ & ${98.31}$ & ${97.25}$ & ${99.64}$ & ${97.25}$ & ${95.56}$ & ${98.65}$ & \textbf{97.65}\\

\hline
Dataset & Method & T-shirt & Trouser & Pullover & Dress & Coat & Sandal & Shirt & Sneaker & Bag & Ankle boot & Mean\\

\hline

FMNIST
& DAGMM\cite{zong2018deep} & 30.3 & 31.1 & 47.5 & 48.1 & 49.9 & 41.3 & 42.0 & 37.4 & 51.8 & 37.8 & 41.7\\
& DSEBM\cite{zhai2016deep} & 89.1 & 56.0 & 86.1 & 90.3 & 88.4 & 85.9 & 78.2 & 98.1 & 86.5 & 96.7 & 85.5\\
& LSA\cite{abati2019latent} & 91.6 & 98.3 & 87.8 & 92.3 & 89.7 & 90.7 & 84.1 & 97.7 & 91.0 & 98.4 & 92.2\\
& DSVDD\cite{ruff2018deep} & 98.2 & 90.3 & 90.7 & 94.2 & 89.4 & 91.8 & 83.4 & 98.8 & 91.9 & 99.0 & 92.8\\
& OCSVM\cite{scholkopf2002learning} & 91.9 & 99.0 & 89.4 & 94.2 & 90.7 & 91.8 & 83.4 & 98.8 & 90.3 & 98.2 & 92.8\\

& \textbf{Ours (ALPS)} & ${94.42}$ & ${98.46}$ & ${89.82}$ & ${90.65}$ & ${91.68}$ & ${90.40}$ & ${80.43}$ & ${97.88}$ & ${97.87}$ & ${97.88}$ & \textbf{92.94}\\

\hline

\end{tabular}}
\end{table*}

\subsection{Approach}
The two modules of our framework, the autoencoder and the adversarial distorter, are trained with two opposing goals. During training, the autoencoder aims to reconstruct perturbed inputs with minimal error, whereas the adversarial distorter aims to increase it by adding perturbations to the latent space. To train the networks, the same input is given to both the autoencoder and adversarial distorter. We add the generated perturbation to the latent space of the encoder. The perturbed latent space is then passed to the decoder for reconstruction. The involvement of perturbations introduces a new task for the autoencoder. In addition to reconstruction, the autoencoder will learn to neutralize the effect of added perturbations during training. During test time, we pass the input to both modules. Since the autoencoder is optimized only on inliers, it is unable to neutralize the perturbations added to the latent space of the outliers. During this procedure, the perturbations allow the autoencoder to learn different variations in the latent space of inliers. When tested with an outlier, it will modify its latent space to be similar to that of inliers, which results in high reconstruction error. To avoid the dominance of the adversarial distorter, we update its weights every $n$ epochs. In addition, we created a weighted loss for both modules, giving larger values to the reconstruction loss for the autoencoder due to the high complexity of its task. 

\begin{table}
\centering
\caption{Frame-level AUCROC and EER comparison \% on UCSD dataset with state-of-the-art methods.}
 \begin{tabular}{|l|l|c|}
 \hline
Method & AUCROC (\%) & EER (\%) \\ [0.5ex] 
 \hline
TSC \cite{luo2017revisit_novelty}                       & 92.2 & -                  \\
FRCN action \cite{hinami2017joint_novelty}               & 92.2                  & -                  \\
AbnormalGAN \cite{ravanbakhsh2017abnormal_novelty}               & 93.5                   & 13               \\
MemAE \cite{gong2019memorizing} & 94.1                   & -                \\
GrowingGas \cite{sun2017online}                & 94.1   & -                \\
FFP \cite{liu2018future_novelty}     & 95.4  & -                  \\
ConvAE+UNet \cite{Nguyen_2019_ICCV}               & 96.2  & -                  \\
STAN \cite{lee2018stan}        & 96.5   &   -    \\
Object-centric \cite{ionescu2019object} &  97.8   & -                       \\ 
Ravanbakhsh \cite{ravanbakhsh2019training} & - & 14       \\
ALOCC \cite{sabokrou2018adversarially} & - & 13       \\
Deep-cascade \cite{sabokrou2017deep_novelty} & - & 9       \\
 Old is gold \cite{zaheer2020old} & 98.1 & 7       \\

 \textbf{Ours (ALPS)}  &  \textbf{98.16} &  \textbf{6} \\
 
 \hline
\end{tabular}

\label{ucsd_exp}
\end{table}

\subsection{Anomaly Scores}
The loss term in ALPS presents an opportunity for three anomaly scores to be defined: the normal reconstruction loss, the perturbed reconstruction loss and the average of them both. Normal reconstruction loss is calculated using mean squared error (MSE) between the input and the reconstruction. In the perturbed reconstruction loss, we calculate MSE while the latent space of the input is perturbed. Both losses are scaled between 0 and 1 to calculate the anomaly scores. The third anomaly score is calculated by averaging between the two losses before scaling. The best score among these three were reported.

\begin{table}
\centering
\caption{Comparison of various autoencoders for anomaly detection on MNIST}
 \begin{tabular}{|l|l|c|}
 \hline
Method & AUCROC (\%)\\ [0.5ex] 
 \hline
Vanilla AE & 95.42    \\
 AE with added random noise & 95.95\\

 \textbf{AE with ALPS}  &  \textbf{97.65} \\
 
 \hline
\end{tabular}

\label{mnist_ex2}
\end{table}

\section{Experiments and Results}
The following section provides a detailed analysis of the proposed method. The results of ALPS are compared to recent and state-of-the-art methods in the literature on three datasets that are benchmarks in anomaly detection. Throughout all experiments, the method is trained exclusively on inlier samples. Moreover, a validation set containing 150 samples from inliers and 150 samples from outliers from the training set is used to determine the optimal epoch for selecting models.

In this paper, we chose MNIST \cite{lecun1998mnist}, FMNIST \cite{xiao2017fashion} and UCSD \cite{chan2008ucsd} for anomaly detection. These benchmark datasets are widely used in the anomaly detection literature. In the following, we provide descriptions of each dataset as well as the protocols for evaluation. 

MNIST is a dataset of handwritten digits with 60,000 grayscale images with a resolution of 28 by 28. This is one of the most popular benchmark datasets in anomaly detection. A similar dataset, FMNIST contains 60,000 images of 28 by 28 grayscale fashion accessories but since there is a substantial amount of intra-class variation, it is more challenging for anomaly detection than MNIST. For these two datasets, the protocol we follow is to designate one class as normal data and other classes as outliers. As a measure of anomaly detection performance, we calculate the Area Under the Curve (AUROC) for each class and report the average for all classes. Table \ref{table:mnist,fnmnist} summarizes the AUROC results for each dataset. The results of our study indicate ALPS is superior to the state-of-the-art autoencoder methods for anomaly detection. Figure \ref{mnist_vis} and \ref{fmnist_vis} shows the visual performance of model on MNIST and FMNIST datasets, respectively. As shown, the abnormal images are reconstructed similar to normal images after adding perturbations.

As part of our method evaluation, we selected the UCSD video dataset. It contains outdoor scenes with pedestrians, cars, skateboards, wheelchairs, and bicycles. A normal frame is defined as only containing pedestrians, whereas an anomalous frame contains other objects. This dataset has two subsets named Ped1 and Ped2. Ped1 contains 34 training videos and 36 testing videos, while Ped2 contains 2,550 frames in 16 training videos and 2,010 frames in 12 testing videos, all of which have a resolution of 240 by 360 pixels. This dataset is evaluated using a patch-based protocol where each frame is divided into 30 by 30 sections. In training, only pedestrians are included in the patches, yet, the model was evaluated on patches that contained pedestrians or other objects at the test time. A frame-level AUROC and Equal Error Rate (EER) are calculated in Table \ref{ucsd_exp} to report the performance of our method on this dataset. ALPS surpasses state-of-the-art methods for detecting video anomalies on UCSD. The visual performance of the model is depicted in Figure \ref{ucsd_vis}.

Furthermore, we demonstrated the effectiveness of the added perturbations in a complementary experiment. Compared to autoencoders trained with only random noise and a normal autoencoder, adversarially learned perturbations can significantly improve autoencoders' performance for anomaly detection. Table \ref{mnist_ex2} shows the result of this experiment on MNIST dataset.

\section{Conclusion}
In this paper, we present an adversarial framework for detecting anomalies in both images and videos. In particular, our method includes a convolutional encoder-decoder (Autoencoder) that tries to reconstruct perturbed images and an encoder (Adversarial Distorter) that attempts to generate effective perturbations from input data. The Adversarial Distorter will increase the reconstruction loss by perturbing the latent space of the input, while the Autoencoder attempts to minimize it. Adding perturbations to the latent space of autoencoders allows them to learn richer representations and improves anomaly detection performance at test time. The method is applicable to a variety of tasks, including the detection of outliers and anomalies in images and videos. The results demonstrate that ALPS outperforms recent state-of-the-art models for identifying anomalies.





%




\bibliography{IEEEfull}
\end{document}